\documentclass[conference]{IEEEtran}

\usepackage{cite}
\usepackage{amsmath,amssymb,amsfonts}
\usepackage{algorithmic}
\usepackage{graphicx}
\usepackage{textcomp}
\usepackage{xcolor}
\usepackage{booktabs}
\usepackage{multirow}
\usepackage[hypertexnames=false]{hyperref}
\usepackage{array}
\usepackage{tabularx}
\usepackage{float}

\def\BibTeX{{\rm B\kern-.05em{\sc i\kern-.025em b}\kern-.08em
    T\kern-.1667em\lower.7ex\hbox{E}\kern-.125emX}}

\begin{document}

\title{Explainable Machine Learning for Broadband Adoption Disparities: Tract-Level Prediction and SHAP-Based Factor Profiling}

\author{\IEEEauthorblockN{Xiao Han}
\IEEEauthorblockA{\textit{Independent Researcher} \\
San Jose, CA, USA \\
xhan@alumni.emory.edu}}

\maketitle

\begin{abstract}
The United States has allocated approximately \$65 billion through the Infrastructure Investment and Jobs Act for broadband expansion, yet evidence-based methods for targeting these investments remain underdeveloped. This paper presents an explainable machine learning framework for profiling broadband adoption disparities at census-tract granularity across 83,359 tracts nationwide. Using 65 socioeconomic, demographic, and infrastructure features derived from the American Community Survey 2022, we train a LightGBM model under spatial five-fold cross-validation, achieving $R^2 = 0.533$ and Spearman $\rho = 0.763$; state-held-out cross-validation (51 folds) confirms generalization ($R^2 = 0.525$). TreeSHAP analysis identifies income and education as the dominant factor group (with the engineered interaction term absorbing attribution from its constituent features), and SHAP-based clustering reveals three exploratory factor profiles: Well-Connected Moderate (${\approx}$49K tracts), Affordability-Limited Severe (${\approx}$21K tracts), and Rural--Elderly (${\approx}$13K tracts). As a screening tool, ML-based tract selection captures 38.0\% of the total adoption gap within the top 10\% of tracts versus 35.2\% for income-only heuristics (+2.8~pp, $p < 0.002$, county-block bootstrap); in regret-reduction terms, the model closes 19\% of the remaining gap between income-only and oracle selection. The primary contribution is the per-tract factor decomposition: SHAP identifies which feature groups (income/education, rurality, age) are most strongly associated with each tract's predicted gap, and informs differentiated investigation. A temporal stability check, training on ACS 2017 and predicting ACS 2022 with zero survey-year overlap, confirms ranking stability ($\rho = 0.784$, noting hyperparameters tuned on 2022 data).
\end{abstract}


\section{Introduction}

The digital divide, the gap between those who have access to and effectively use digital technologies and those who do not, remains one of the most persistent equity challenges in the United States. Despite decades of policy attention, approximately 24 million Americans lack access to broadband internet, and tens of millions more who live in areas with nominal availability do not subscribe due to affordability, digital literacy, or relevance barriers~\cite{lai2021revisiting, horrigan2010broadband}. The COVID-19 pandemic exposed the severity of this divide as essential services moved online, transforming broadband from a convenience to a necessity~\cite{lai2021revisiting}.

In response, the Infrastructure Investment and Jobs Act (IIJA) of 2021 allocated approximately \$65 billion for broadband, including \$42.45 billion for the Broadband Equity, Access, and Deployment (BEAD) program, the largest broadband investment in U.S. history~\cite{iija2021}. BEAD allocates funding based on each state's share of unserved and high-cost unserved locations as identified through the FCC's broadband maps~\cite{ntia2023bead}. However, this approach relies on a small number of heuristic factors and focuses primarily on infrastructure availability rather than the full range of demand-side barriers that contribute to adoption disparities. Implementation challenges have compounded the problem: one state's verified eligible location list contained only 22.6\% of the originally mapped unserved locations~\cite{crs2025bead}.

Machine learning offers the potential to identify complex, non-linear relationships among the many factors associated with broadband adoption disparities. However, to be useful for public policy, such models must be not only accurate but also interpretable. Policymakers need to understand \textit{why} specific areas have low adoption, whether due to affordability constraints, rurality, age-related barriers, or combinations of these, to investigate appropriate interventions. Recent advances in explainable artificial intelligence (XAI), particularly TreeSHAP~\cite{lundberg2020local}, enable faithful decomposition of gradient-boosted model predictions into per-feature contributions grounded in cooperative game theory.

This paper connects digital divide research, machine learning, and policy action by presenting an integrated framework that (1)~predicts broadband adoption disparities at census-tract granularity, (2)~explains predictions using SHAP to identify which factor groups contribute most to each tract's predicted gap, and (3)~clusters tracts into exploratory factor profiles. The novelty lies not in any individual algorithmic component (LightGBM, TreeSHAP, and $k$-means are established methods) but in the \textit{integrated framework} that combines prediction, explanation, and profiling at national scale for broadband policy, together with methodological care in leaky-feature exclusion, spatial cross-validation, and policy-category-aware SHAP aggregation. Specifically, our contributions are:

\begin{itemize}
    \item \textbf{Tract-level ML prediction at national scale:} We train LightGBM on 83,359 census tracts with 65 features under spatial five-fold cross-validation, achieving $R^2 = 0.533$ and Spearman $\rho = 0.763$; state-held-out CV (51 folds) confirms generalization at $R^2 = 0.525$.
    \item \textbf{SHAP-based factor profiling:} TreeSHAP decomposes predicted adoption gaps into feature-level contributions, revealing that income and education features collectively carry the largest attribution. SHAP-based clustering identifies three exploratory factor profiles with distinct associated patterns, enabling differentiated investigation.
    \item \textbf{ML-based policy screening:} ML-based tract screening captures 38.0\% of the total national adoption gap within the top 10\% of tracts versus 35.2\% for income-only heuristics ($+2.8$~pp, $p < 0.002$). The primary contribution is the per-tract factor decomposition that identifies which feature groups are most strongly associated with each tract's predicted gap.
\end{itemize}

The remainder of this paper is organized as follows. Section~II reviews related work. Section~III describes our data and feature engineering. Section~IV details the methodology. Section~V presents results. Section~VI discusses findings, implications, and limitations. Section~VII concludes.

\section{Related Work}

\subsection{The Digital Divide and Broadband Adoption}

Research on the digital divide has evolved through successive phases. Van Dijk~\cite{vandijk2006digital} proposed a resource and appropriation theory arguing that categorical inequalities produce unequal ICT access, and later~\cite{vandijk2020digital} demonstrated that digital inequality reinforces rather than reduces existing social inequality. Blank and Groselj~\cite{blank2014dimensions} formalized the multi-dimensional view, proposing that internet use should be measured across amount, variety, and types of activities.

A substantial body of work has documented the geography of broadband disparities. Grubesic~\cite{grubesic2006spatial} developed a spatial taxonomy of broadband deployment regions, while a subsequent analysis~\cite{grubesic2012broadband} found systematic overstatement of coverage. Salemink et al.~\cite{salemink2017rural} reviewed 157 papers on the rural digital divide, identifying persistent infrastructure quality gaps. Horrigan~\cite{horrigan2010broadband} identified three primary barriers (cost, digital literacy, and perceived irrelevance) and categorized non-adopters into four segments.

Prieger~\cite{prieger2013broadband} examined both fixed and mobile broadband access in rural areas, finding that mobile partially compensates for fixed-line deficits but income-based disparities persist. Perrin and Turner~\cite{perrin2019smartphones} reported that Black and Hispanic adults remain significantly less likely to subscribe to home broadband, with approximately one-quarter relying on smartphones as their sole connection.

The distinction between availability and adoption is central to this literature. Whitacre et al.~\cite{whitacre2014broadband, whitacre2014rural} provided causal evidence that broadband \textit{adoption}, not mere availability, positively affects firm counts, employment, and median household income. Whitacre et al.~\cite{whitacre2015infrastructure} further decomposed the metro--non-metro adoption gap, finding that infrastructure explains part of the gap but socioeconomic factors account for a significant additional portion. Gallardo~\cite{gallardo2022ddi} developed the Digital Divide Index at county level, but without predictive modeling or feature-level explanation. Despite this extensive literature, no prior work has applied modern machine learning to predict broadband adoption disparities at census-tract granularity at national scale or decomposed predicted factors into actionable policy categories.

\subsection{Machine Learning for Broadband and Telecom Prediction}

Machine learning has been applied to various aspects of telecommunications but rarely to broadband adoption prediction. Oughton and Mathur~\cite{oughton2021predicting} used convolutional neural networks with satellite imagery to predict cell phone adoption in Malawi and Ethiopia, outperforming baselines by over 40\% in variance explained; however, their work targeted developing countries and cell phone rather than broadband adoption. Singleton et al.~\cite{singleton2020mapping} integrated XGBoost into a small-area estimation framework to map digital inequality across Great Britain but focused on internet usage patterns rather than broadband subscription.

Zahnd et al.~\cite{zahnd2022geographic} used spatial regression to examine broadband access disparities across U.S. census tracts, finding that poverty and education are the strongest predictors. While this work operates at our target spatial resolution, it relies on spatial regression rather than ML and does not produce prediction models or explainability-driven recommendations. Paul et al.~\cite{paul2023decoding} analyzed broadband plan disparities across 837,000+ addresses in 30 cities, finding ISP pricing varies up to 600\% within a city, a supply-side analysis that complements our demand-side prediction. Nabi et al.~\cite{nabi2024red} applied XGBoost with SHAP to identify suspicious ISP availability claims in the FCC's National Broadband Map (AUC $> 0.98$), demonstrating feasibility but targeting data quality auditing rather than adoption prediction. Agarwal and Canfield~\cite{agarwal2024rural} proposed a theory-driven agent-based model to simulate broadband adoption dynamics in a single Missouri community of 274 households, the most recent adoption prediction attempt but limited in scale. Together, these studies show that no prior work has applied gradient-boosted models to predict broadband adoption at the census-tract level across the entire United States.

\subsection{Explainable AI and SHAP for Policy Applications}

Lundberg and Lee~\cite{lundberg2017unified} introduced SHAP, a unified framework grounded in cooperative game theory. Lundberg et al.~\cite{lundberg2020local} extended this with TreeSHAP, a polynomial-time algorithm for exact SHAP computation on tree ensembles. Rudin~\cite{rudin2019stop} argued that post-hoc explanations are unreliable for high-stakes individual decisions; in our setting, SHAP informs aggregate policy screening where Shapley-value guarantees provide sufficient faithfulness. Athey and Imbens~\cite{athey2019machine} cautioned that feature importance is predictive rather than causal, a distinction we maintain throughout.

SHAP has been deployed in policy-relevant domains. Lundberg et al.~\cite{lundberg2018explainable} applied SHAP interaction values to predict surgical risk, demonstrating that SHAP can shift focus from prediction to prevention by identifying modifiable risk factors, an approach we adapt for broadband policy. Wagner et al.~\cite{wagner2022shap} applied SHAP to gradient-boosted models of 3.5 million commutes, translating outputs into spatial policy recommendations, the closest methodological analog to our work.

\subsection{Federal Broadband Investment}

The IIJA~\cite{iija2021} allocated approximately \$65 billion for broadband, including \$42.45 billion for BEAD~\cite{ntia2023bead}, \$2.75 billion for the Digital Equity Act~\cite{digitalequityact2021}, and \$14.2 billion for the Affordable Connectivity Program (ACP). The ACP enrolled approximately 23 million households before ending in June 2024; its termination is a major affordability shock. BEAD's June 2025 restructuring shifted the program toward lowest-cost deployment, reducing alignment with demand-side analysis. A Congressional Research Service report~\cite{crs2025bead} documented mapping inaccuracies and implementation delays. This policy context motivates tools that can predict where adoption disparities are most severe and characterize which factors are associated with them.

\section{Data and Feature Engineering}

\subsection{Data Sources and Target Variable}

Our primary data source is the American Community Survey (ACS) 2022 five-year estimates, which provides socioeconomic, demographic, and housing characteristics at the census-tract level. The ACS five-year estimates average data collected from 2018 through 2022, spanning the COVID-19 pandemic; this temporal blending should be considered when interpreting results. We complement ACS data with County Business Patterns (CBP) data for economic activity indicators. CBP features are available only at county resolution and are shared across all tracts within a county, limiting within-county discrimination but providing useful between-county economic context. The final dataset encompasses 83,359 census tracts (${\sim}$98\% of all U.S. tracts); the ${\sim}$1,700 excluded tracts lack sufficient ACS responses (primarily very-low-population tracts, including some tribal/reservation areas) and may disproportionately represent underserved populations.

The target variable is the \textit{relative} broadband adoption gap: $g_i = (m - r_i) / m$, where $m$ is the national median broadband subscription rate and $r_i$ is tract $i$'s rate. Positive values indicate underserved tracts. Critically, we exclude 23 internet self-report features (e.g., ``has any internet subscription'') to avoid predicting the outcome from near-proxies of itself.

\subsection{Feature Engineering}

We construct 65 features organized into ten categories: \textbf{Income/Poverty (6):} log median household income, median household income, income per capita, poverty rate, Gini index, and home value-to-income ratio. \textbf{Education (4):} percentage with less than high school, high school diploma only, bachelor's degree or higher, and education gap. \textbf{Demographics/Race (6):} percentage White, Black, Asian, Hispanic, and American Indian/Alaska Native, plus racial diversity index. \textbf{Age (4):} percentage under 18, percentage 65+, median age, and age dependency ratio. \textbf{Housing (7):} occupancy rate, median home value, log median home value, owner-occupied and renter-occupied percentages, persons per housing unit, and total housing units. \textbf{Employment (4):} labor force participation rate, unemployment rate, tech wage premium, and tech employment share. \textbf{Business Environment (10):} CBP-derived county-level features including total and tech-sector establishments, employment, and payroll. \textbf{Urbanicity/Density (8):} urbanicity indicators, land area, population density, and housing density. \textbf{Population (2):} total and log population. \textbf{Interaction Terms (14):} cross-domain interaction terms including income$\times$education, elderly$\times$renter, rural$\times$elderly, and rural$\times$poverty.

\section{Methodology}

\subsection{Problem Formulation}

Let $\mathbf{X} \in \mathbb{R}^{n \times p}$ denote the feature matrix with $n = 83{,}359$ tracts and $p = 65$ features, and let $\mathbf{y} \in \mathbb{R}^n$ denote the broadband adoption gap vector. We learn a mapping $f: \mathbb{R}^p \rightarrow \mathbb{R}$ that predicts the adoption gap $\hat{y}_i = f(\mathbf{x}_i)$ for each tract, evaluated using RMSE, $R^2$, MAE, and Spearman $\rho$.

\subsection{LightGBM with Spatial Cross-Validation}

We select LightGBM~\cite{ke2017lightgbm} for its efficiency, native missing-value handling, and compatibility with TreeSHAP for exact Shapley value computation. Hyperparameters are optimized using Optuna~\cite{akiba2019optuna} with 50 Bayesian optimization trials. Evaluation uses spatial five-fold cross-validation, partitioning tracts into five geographic regions (Northeast, Southeast, Midwest, Southwest, West) to prevent spatial information leakage~\cite{roberts2017cross}. The optimized hyperparameters are: number of leaves = 281, maximum depth = 7, learning rate = 0.016, and number of estimators = 1,141.

We compare against seven baselines: XGBoost~\cite{chen2016xgboost}, CatBoost, Random Forest, Ridge, Lasso, MLP, and OLS, all evaluated under identical spatial five-fold CV. Gradient-boosted baselines use default hyperparameters; this is not a controlled tuning comparison but demonstrates that LightGBM's tuned performance is achievable at or near default settings across the gradient-boosted family.

\subsection{SHAP Analysis and Clustering}

TreeSHAP~\cite{lundberg2020local} decomposes predictions into additive feature contributions: $f(\mathbf{x}_i) = \phi_0 + \sum_{j=1}^{p} \phi_j(\mathbf{x}_i)$, where $\phi_0$ is the expected prediction and $\phi_j(\mathbf{x}_i)$ is the SHAP value of feature $j$ for tract $i$. SHAP values are computed from a model trained on the full dataset, on a stratified sample of 20,000 tracts. SHAP interaction values are computed on a 2,000-tract subsample due to $O(p^2)$ cost.

We apply $k$-means clustering on the SHAP value matrix to identify factor profiles, grouping tracts by \textit{why} they are predicted to have gaps rather than by demographic similarity. Silhouette scores favor $k{=}2$ (0.40) over $k{=}3$ (0.35), but $k{=}2$ yields only a binary rural/non-rural split. We select $k{=}3$, which further subdivides the non-rural group by poverty and education, as the smallest $k$ that separates interpretable profiles. This is a pragmatic, exploratory choice; $k{=}3$ is still substantially a rurality partition. Gaussian mixture modeling ($k{=}3$) yields high within-method confidence but substantially different assignments (ARI = 0.290 vs.\ $k$-means), confirming that the specific three-group structure is method-dependent.

\subsection{Policy Screening}

We compare four strategies for selecting the top 10\% of tracts ($n = 8{,}335$) for prioritized investigation: (1)~Gap-Ranked (oracle upper bound using actual gaps), (2)~Model-Predicted (ranking by out-of-fold predicted gaps), (3)~Income-Only (ranking by lowest median household income), and (4)~Random baseline. Gap-capture@10\% is the sum of selected tracts' positive gaps divided by all positive gaps nationally ($\sum_{i: g_i > 0} g_i = 4{,}132$), with 95\% CIs from county-level block bootstrap (500 resamples, 3,114 counties).

\subsection{Temporal Stability Check}

To demonstrate temporal generalization, we train on ACS 2017 five-year estimates (2013--2017) and predict ACS 2022 (2018--2022), achieving zero survey-year overlap. Because 2017 estimates use 2010 census tract boundaries while 2022 uses 2020 boundaries, we apply the Census Bureau's 2010$\to$2020 tract relationship file, converting raw ACS count variables via area-weighted aggregation and recomputing rate features from the converted counts. Of 73,782 source tracts, 58.1\% map one-to-one; the remainder are split or merged. After crosswalk conversion, 82,446 tracts are common to both vintages. We train LightGBM with the same hyperparameters tuned on the 2022 spatial CV folds. Because hyperparameters were selected on the evaluation vintage, this is a temporal stability check rather than a fully external validation; results may be optimistic relative to tuning on 2017 data alone.

\section{Experimental Results}

\subsection{Model Comparison}

Table~\ref{tab:model_comparison} presents model performance under spatial five-fold CV. LightGBM achieves $R^2 = 0.533 \pm 0.063$ and Spearman $\rho = 0.763 \pm 0.037$ ($\pm$ values are fold standard deviations, $n = 5$).

\begin{table}[t]
\centering
\caption{Model Performance Under Spatial 5-Fold Cross-Validation}
\label{tab:model_comparison}
\begin{tabular}{lcccc}
\toprule
\textbf{Model} & \textbf{RMSE} & \textbf{MAE} & \textbf{$R^2$} & \textbf{Spearman $\rho$} \\
\midrule
LightGBM & 0.069 & 0.049 & 0.533 & 0.763 \\
XGBoost & 0.069 & 0.049 & 0.530 & 0.762 \\
CatBoost & 0.069 & 0.049 & 0.529 & 0.762 \\
Random Forest & 0.071 & 0.050 & 0.515 & 0.759 \\
Ridge & 0.079 & 0.053 & 0.380 & 0.733 \\
Lasso & 0.090 & 0.060 & 0.153 & 0.696 \\
MLP & 0.087 & 0.058 & 0.059 & 0.717 \\
OLS & 0.106 & 0.070 & $-0.311$ & 0.626 \\
\bottomrule
\end{tabular}
\end{table}

The three gradient-boosted models perform similarly at the fold level ($R^2$ ranges: 0.527--0.540; with only five folds, formal significance tests lack power). We select LightGBM for native TreeSHAP compatibility. The gap between non-linear ensembles ($R^2 \approx 0.53$) and linear models ($R^2 \leq 0.38$) demonstrates that non-linear feature interactions are important for this prediction task. MLP exhibits high variance ($R^2$ range: $-1.73$ to $0.58$), reflecting sensitivity to distribution shift under spatial CV. OLS produces negative $R^2$, indicating overfitting to spatial patterns.

Computing $R^2$ on all 83,359 concatenated out-of-fold predictions yields $0.568$; we use the conservative per-fold mean ($0.533$) as the headline.

\textbf{Tuning--evaluation separation.} Hyperparameters and evaluation share the same spatial CV folds. To bound the resulting optimism, we conduct state-held-out CV (51 folds: 50 states + DC; Puerto Rico and territories are excluded from the ACS tract-level dataset). Tuned LightGBM achieves mean $R^2 = 0.525 \pm 0.105$ (weighted $R^2 = 0.531$, $\rho = 0.737 \pm 0.061$). Default-hyperparameter LightGBM achieves $R^2 = 0.518$; tuning yields only 0.007~$R^2$ improvement, suggesting limited scope for tuning-induced optimism.

\textbf{Model provenance for downstream analyses.} SHAP values are computed from a model trained on the full dataset to characterize global feature attributions. The policy screening framework uses out-of-fold predictions, avoiding data leakage in screening evaluation. The top-10 features are identical across all five fold-specific models, supporting the stability of the full-data attributions.

\textbf{Performance on underserved tracts.} On the policy-relevant subset (gap $> 0$, $n = 41{,}274$), pooled $R^2 = 0.205$ and Spearman $\rho = 0.572$ (mean fold $R^2 = 0.136 \pm 0.105$, mean fold $\rho = 0.542 \pm 0.054$). The lower $R^2$ reflects the reduced target variance within this subset, but the rank correlation confirms meaningful discrimination, the capability exploited by the screening framework, which depends on ranking rather than point-estimate accuracy.

\subsection{SHAP Feature Importance}

\begin{figure}[t]
    \centering
    \includegraphics[width=\columnwidth]{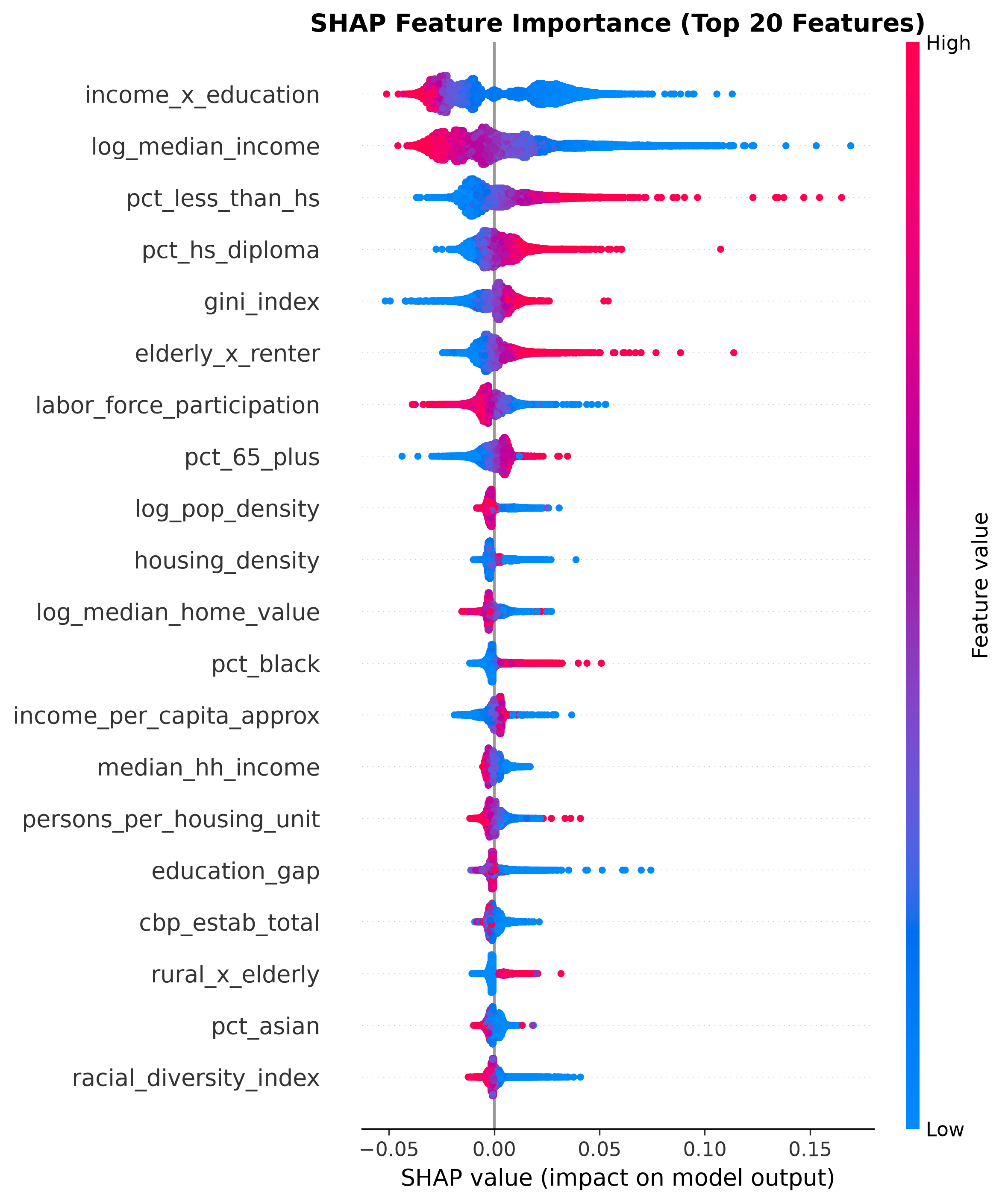}
    \caption{SHAP beeswarm plot. Each dot is one tract; color indicates feature value (red = high, blue = low). Income and education features collectively form the dominant factor group.}
    \label{fig:shap_beeswarm}
\end{figure}

Table~\ref{tab:feature_importance} lists the top 15 features by mean absolute SHAP value, and Fig.~\ref{fig:shap_beeswarm} shows the SHAP beeswarm plot.

\begin{table}[t]
\centering
\caption{Top 15 Features by Mean Absolute SHAP Value}
\label{tab:feature_importance}
\begin{tabular}{clc}
\toprule
\textbf{Rank} & \textbf{Feature} & \textbf{Mean $|\text{SHAP}|$} \\
\midrule
1 & income\_x\_education & 0.0230 \\
2 & log\_median\_income & 0.0182 \\
3 & pct\_less\_than\_hs & 0.0110 \\
4 & pct\_hs\_diploma & 0.0068 \\
5 & gini\_index & 0.0060 \\
6 & elderly\_x\_renter & 0.0056 \\
7 & labor\_force\_participation & 0.0050 \\
8 & pct\_65\_plus & 0.0044 \\
9 & log\_pop\_density & 0.0030 \\
10 & housing\_density & 0.0029 \\
11 & log\_median\_home\_value & 0.0029 \\
12 & pct\_black & 0.0028 \\
13 & income\_per\_capita\_approx & 0.0028 \\
14 & median\_hh\_income & 0.0026 \\
15 & persons\_per\_housing\_unit & 0.0026 \\
\bottomrule
\end{tabular}
\end{table}

Income and education features collectively form the dominant factor group: the top four features by mean $|\text{SHAP}|$ all involve income or education. Tracts with both low income \textit{and} low education are associated with compounding digital exclusion that neither factor alone identifies. Note that the engineered \texttt{income\_x\_education} feature (the product of log median income and bachelor's-degree rate) absorbs SHAP attribution that would otherwise spread across its constituent features individually; its rank-1 position reflects that the tree model concentrates the joint effect into this single feature when available, not that the interaction is uniquely informative beyond its constituents.

When all 14 engineered interaction terms are removed and the model is retrained, performance is unchanged ($R^2 = 0.534$, $\rho = 0.763$) and log\_median\_income (mean $|\text{SHAP}| = 0.030$) and pct\_less\_than\_hs (0.010) become the top two features, absorbing the redistributed SHAP mass. Native SHAP interaction analysis on the ablated model confirms that income$\times$education pairs remain the strongest learned interactions, supporting the substantive finding regardless of whether the interaction is pre-engineered or learned by tree structure.

\begin{figure}[t]
    \centering
    \includegraphics[width=\columnwidth]{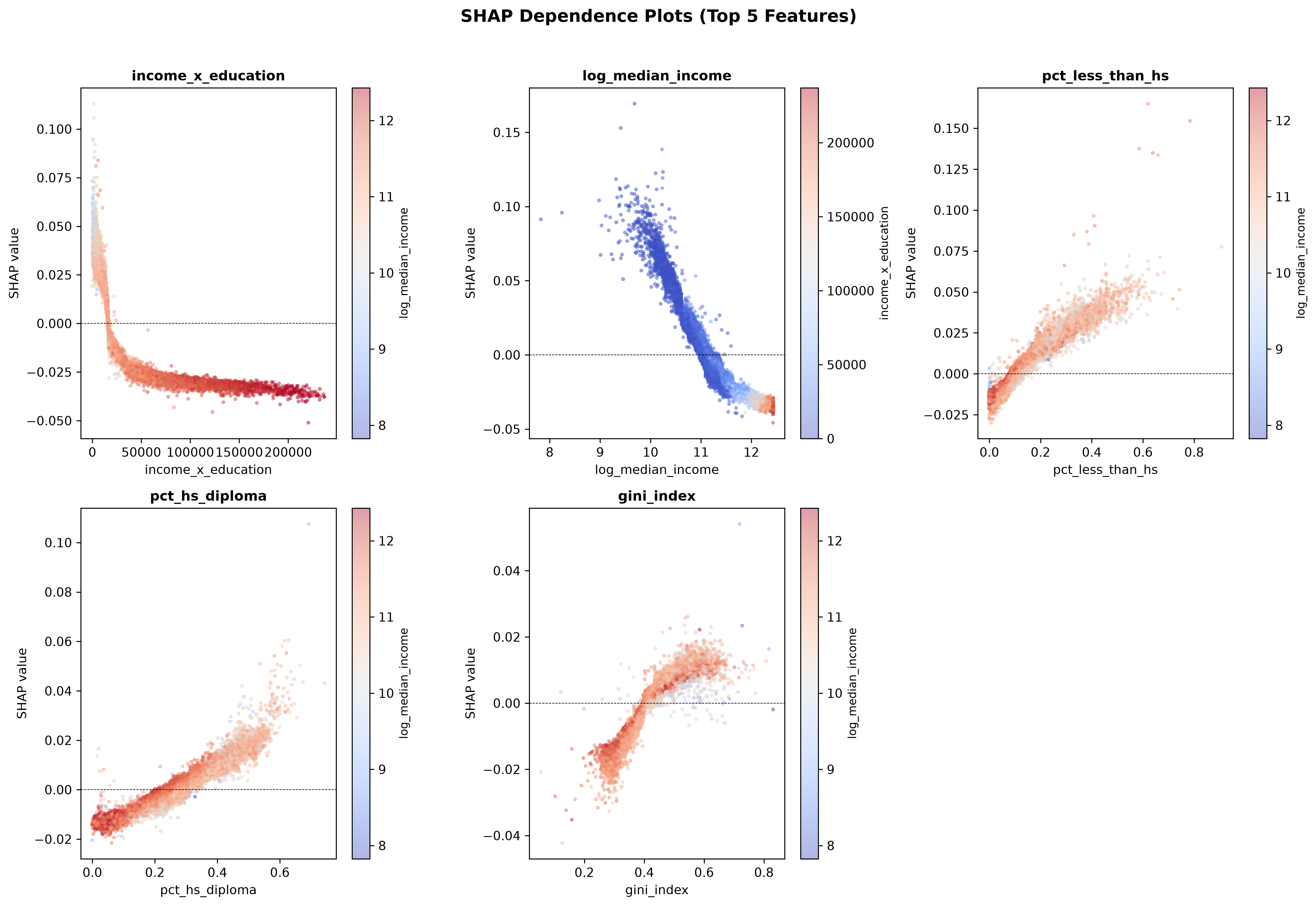}
    \caption{SHAP dependence plots for top features. Non-linear thresholds are visible, particularly for income (sharp increase below \$35K) and education (acceleration above 20\% without high school diploma).}
    \label{fig:shap_dependence}
\end{figure}

Fig.~\ref{fig:shap_dependence} reveals non-linear threshold effects. Log median income below approximately \$35,000 shows a sharp increase in SHAP values, indicating a threshold below which predicted adoption gaps accelerate, consistent with affordability-barrier research~\cite{horrigan2010broadband}.

\subsection{Feature Interactions}

Table~\ref{tab:interactions} presents the top SHAP interaction pairs. The strongest interaction (0.0044) is between income\_x\_education and log\_median\_income, indicating that the joint effect of income and education on predicted adoption gaps varies further depending on absolute income levels.

\begin{table}[t]
\centering
\caption{Top 10 SHAP Feature Interaction Pairs}
\label{tab:interactions}
\begin{tabular}{llc}
\toprule
\textbf{Feature 1} & \textbf{Feature 2} & \textbf{Mean $|\text{Int.}|$} \\
\midrule
income\_x\_education & log\_median\_income & 0.0044 \\
labor\_force\_part. & log\_median\_income & 0.0017 \\
income\_x\_education & labor\_force\_part. & 0.0012 \\
log\_median\_income & pct\_less\_than\_hs & 0.0008 \\
income\_x\_education & pct\_less\_than\_hs & 0.0007 \\
log\_median\_income & pct\_hs\_diploma & 0.0006 \\
income\_per\_capita & income\_x\_education & 0.0006 \\
elderly\_x\_renter & log\_median\_income & 0.0006 \\
education\_gap & log\_median\_income & 0.0005 \\
log\_median\_income & log\_pop\_density & 0.0005 \\
\bottomrule
\end{tabular}
\end{table}

The labor force participation--income interaction (rank 2) is consistent with the possibility that employment provides both digital skills exposure and financial resources for subscriptions. The elderly--renter interaction (rank 8, with log income) identifies a population where age and housing instability co-occur with low adoption.

\subsection{Exploratory Factor Profiles}

\begin{figure}[t]
    \centering
    \includegraphics[width=\columnwidth]{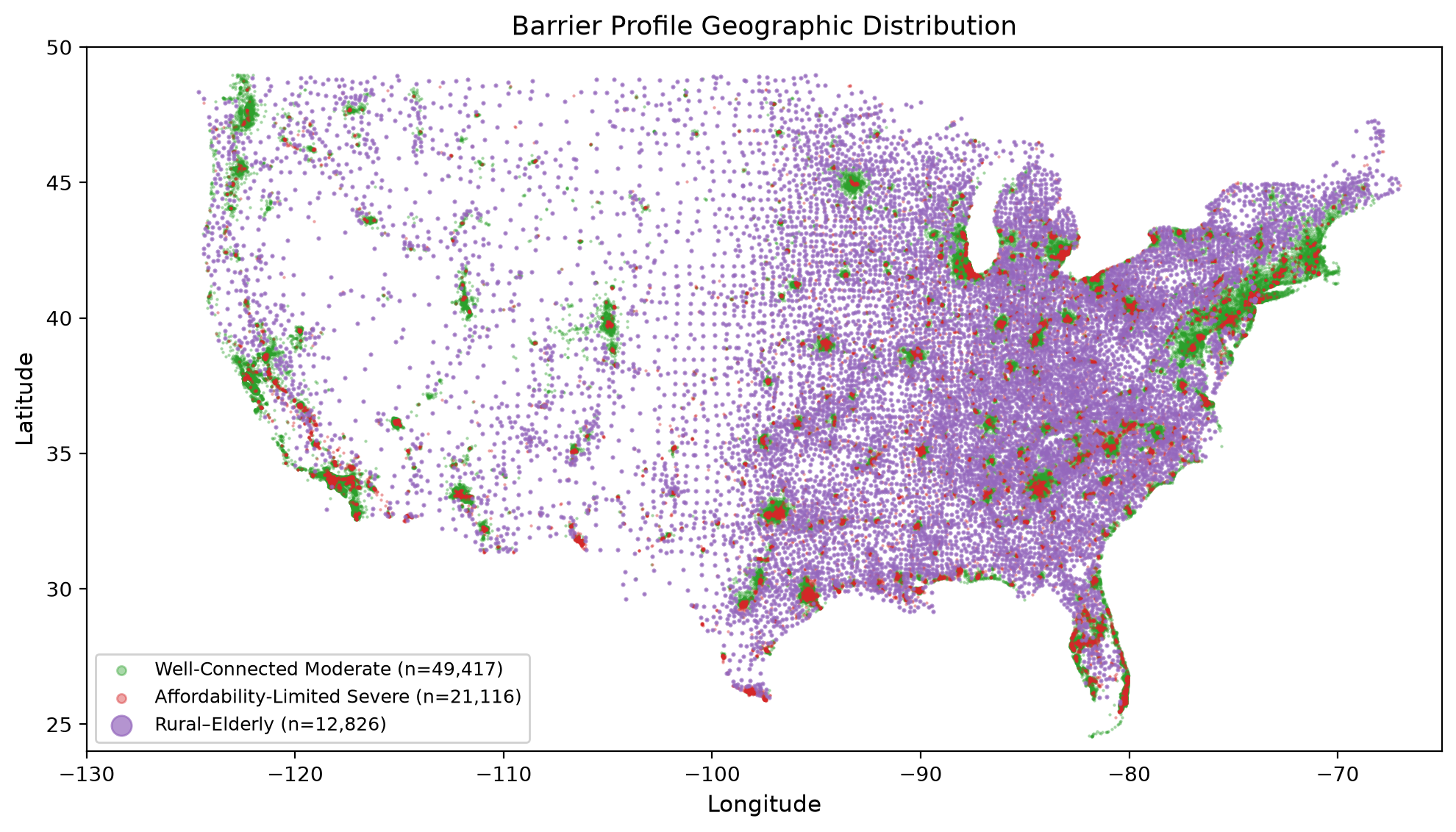}
    \caption{Geographic distribution of exploratory factor profiles. Affordability-Limited Severe tracts concentrate in the Deep South; Rural--Elderly tracts dominate Appalachia and the rural Mountain West.}
    \label{fig:archetype_map}
\end{figure}

To assign all 83,359 tracts to profiles, we compute SHAP values for every tract using LightGBM's native feature-contribution output and predict cluster labels from the $k$-means model fitted on the 20,000-tract sample. Bootstrap stability of the 20,000-tract centroid estimation is high (mean ARI = 0.991 across 50 resamples). Cross-region stability varies: independent $k$-means on each of five spatial regions yields ARI = 0.925 (Southeast) to 0.449 (West), with the West's lower agreement likely reflecting its geographic heterogeneity. The resulting profiles (Table~\ref{tab:archetypes}, Fig.~\ref{fig:archetype_map}) should be treated as an exploratory, policy-oriented taxonomy rather than a uniquely supported clustering result.

\begin{table}[t]
\centering
\caption{Exploratory Factor Profile Characteristics}
\label{tab:archetypes}
\begin{tabular}{lccc}
\toprule
\textbf{Metric} & \textbf{Moderate} & \textbf{Severe} & \textbf{Rural--Elderly} \\
\midrule
$n$ tracts & 49,417 & 21,116 & 12,826 \\
Mean adoption gap & $-0.034$ & 0.095 & 0.106 \\
Poverty rate & 8.5\% & 25.2\% & 13.8\% \\
\% $<$ High school & 7.4\% & 20.0\% & 12.3\% \\
\% Bachelor's+ & 43.3\% & 17.2\% & 19.4\% \\
\% Age 65+ & 17.4\% & 14.8\% & 21.1\% \\
\% Rural & 7.6\% & 3.6\% & 99.3\% \\
Unemployment & 4.7\% & 8.2\% & 5.1\% \\
\bottomrule
\end{tabular}
\end{table}

\textbf{Profile 1: Well-Connected Moderate} (49,417 tracts) has below-average gaps ($-0.034$), high educational attainment (43\% bachelor's+), and low poverty, the connected majority where targeted intervention is less urgent.

\textbf{Profile 2: Affordability-Limited Severe} (21,116 tracts) has high adoption gaps (0.095), characterized by high poverty (25.2\%), low education (20.0\% without high school diploma), and predominantly urban/suburban settings (96.4\% non-rural). SHAP indicates income and education features contribute most to the predicted gap. Concentrated in the Deep South and inner-city areas. The expiration of the Affordable Connectivity Program makes this cohort particularly relevant for policy attention.

\textbf{Profile 3: Rural--Elderly} (12,826 tracts) has the highest gaps (0.106) with a fundamentally different factor pattern: 99.3\% rural, elevated elderly population (21.1\%). SHAP indicates rurality and age-related features contribute most. Concentrated in the Mountain West, Great Plains, and Appalachia. The model contains no infrastructure supply-side data (e.g., FCC broadband availability); rurality serves as a proxy for a bundle of factors including provider competition, terrain, and infrastructure cost that this analysis cannot separate.

The three-profile structure closely mirrors a rurality-then-poverty partition, consistent with $k$-means on the SHAP matrix being dominated by the highest-magnitude features, but the SHAP basis reveals which specific factors contribute most to each profile's predicted gaps. Both are predictive associations; causal validation requires experimental designs.

\subsection{Geographic Distribution}

\begin{figure}[t]
    \centering
    \includegraphics[width=\columnwidth]{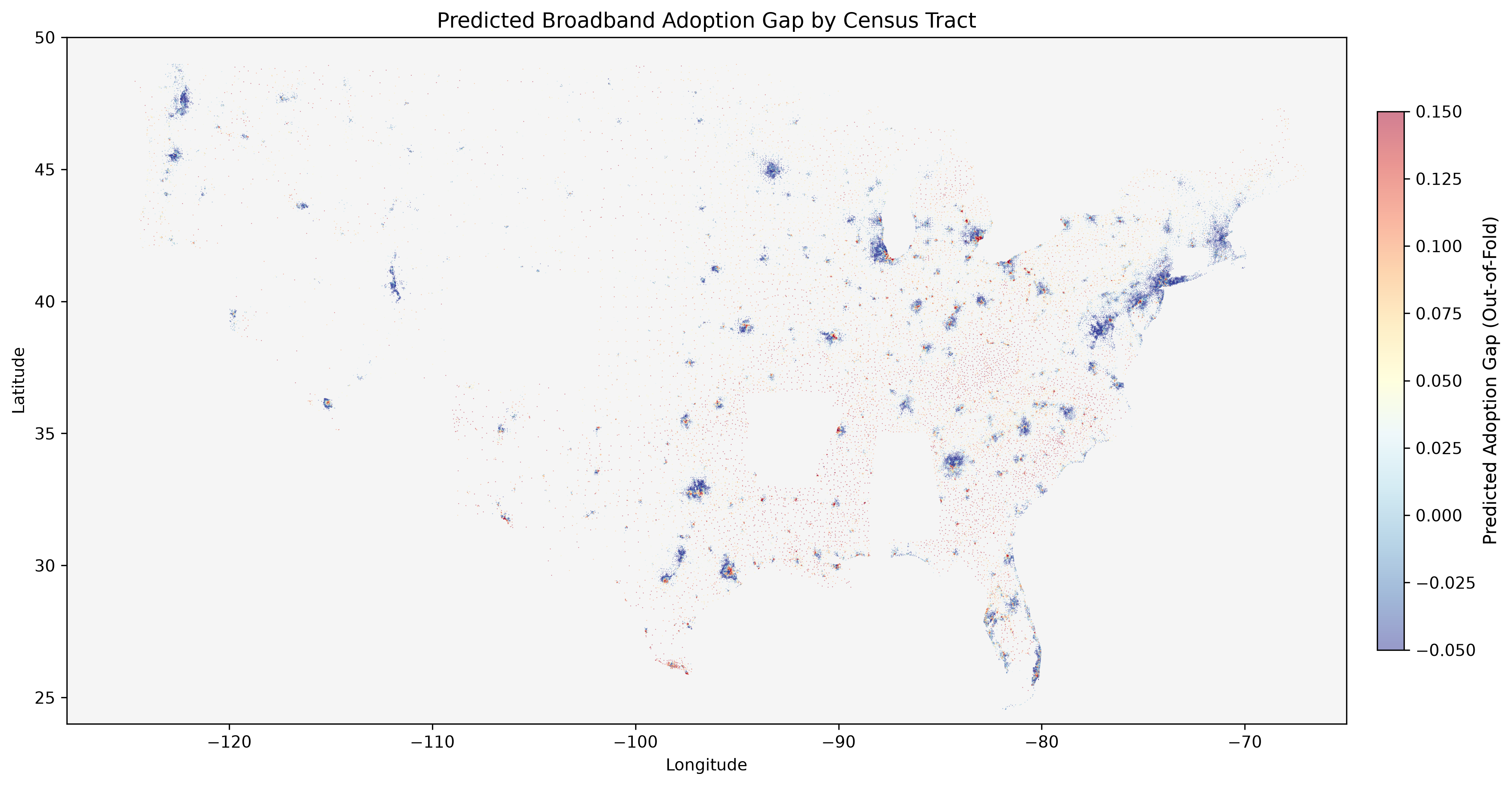}
    \caption{Geographic distribution of predicted broadband adoption gaps (out-of-fold predictions). Strong spatial clustering is visible in the Deep South, Appalachia, and rural Mountain West.}
    \label{fig:geographic_map}
\end{figure}

Fig.~\ref{fig:geographic_map} presents the geographic distribution of predicted adoption gaps from out-of-fold predictions. The highest predicted gaps concentrate in the Deep South (Mississippi, Alabama, Louisiana, primarily Profile~2), Appalachia (West Virginia, eastern Kentucky, a mix of Profiles~2 and~3), and the rural Mountain West (Montana, Wyoming, predominantly Profile~3). The strong spatial clustering validates the spatial CV design.

\subsection{Policy Screening}

\begin{figure}[t]
    \centering
    \includegraphics[width=\columnwidth]{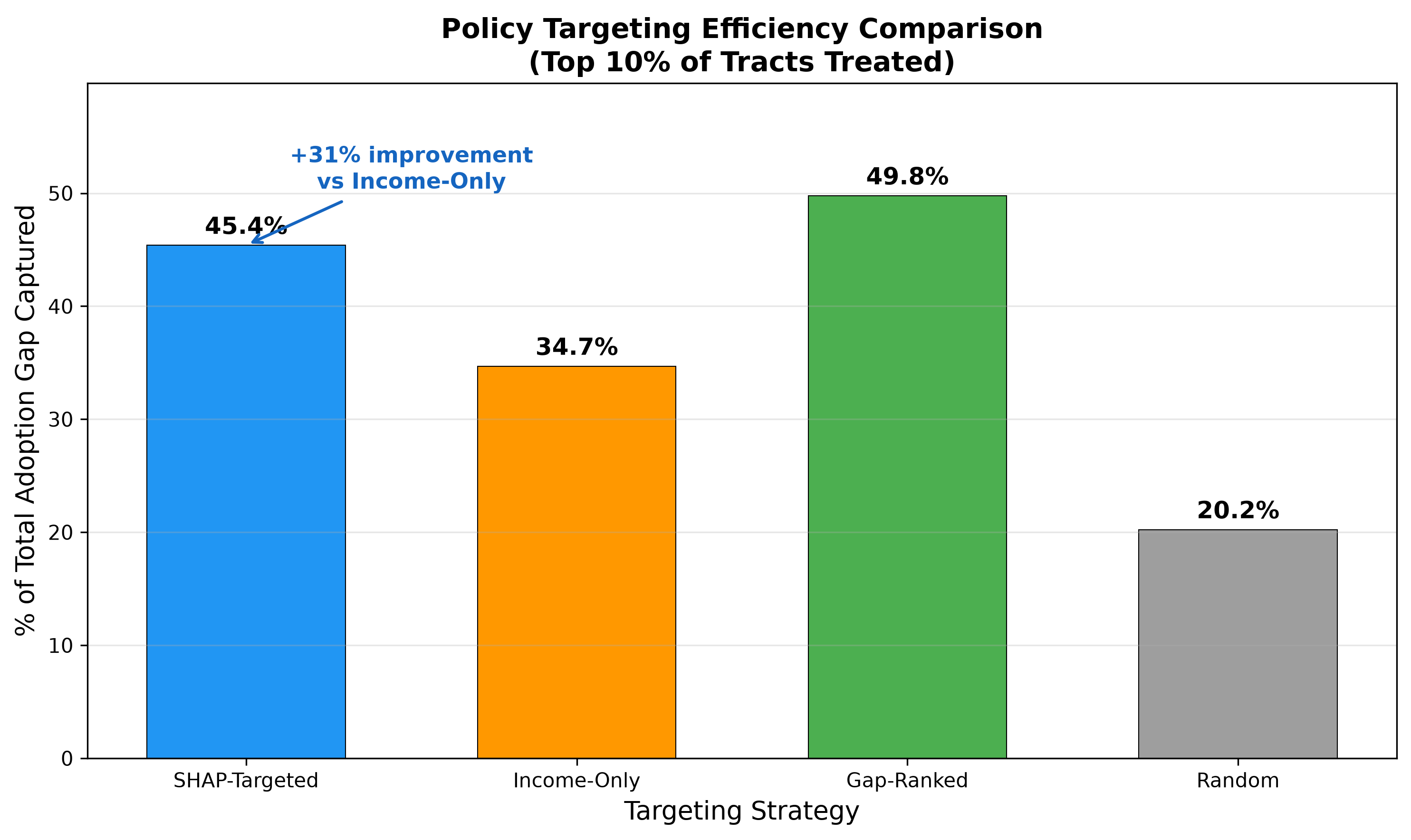}
    \caption{Policy screening comparison. ML-based selection captures 38.0\% of the national adoption gap within the top 10\% of tracts versus 35.2\% for income-only heuristics.}
    \label{fig:policy_comparison}
\end{figure}

Table~\ref{tab:policy} presents the policy screening comparison using out-of-fold predictions and county-level block bootstrap, and Fig.~\ref{fig:policy_comparison} visualizes the results.

\begin{table}[t]
\centering
\caption{Policy Screening Comparison (top 8,335 tracts = 10\%). County-block bootstrap (500 resamples, 3,114 counties).}
\label{tab:policy}
\begin{tabular}{lcc}
\toprule
\textbf{Strategy} & \textbf{Gap Covered} & \textbf{\% Total} \\
\midrule
Gap-Ranked (oracle) & 2,057.8 & 49.8\% \\
Model-Predicted & 1,571.0 & 38.0\% \\
Income-Only & 1,455.2 & 35.2\% \\
Random & 836.5 & 20.2\% \\
\midrule
\multicolumn{3}{l}{\small MP $-$ Income: $+2.8$~pp [2.3, 3.7], $p < 0.002$} \\
\bottomrule
\end{tabular}
\end{table}

ML-based screening captures 38.0\% of the national adoption gap versus 35.2\% for income-only, a gain of $+2.8$~pp (95\% CI [+2.3, +3.7], $p < 0.002$). In regret-reduction terms, income-only recovers 51\% of the oracle headroom above random; the model recovers 60\%, closing 19\% of the remaining gap. The screening improvement arises because income alone misses moderate-income tracts where education or rurality factors are associated with the predicted gap. The primary contribution, however, is not the screening gain but the per-tract factor decomposition: SHAP identifies which feature groups are most strongly associated with each tract's predicted gap.

\begin{figure}[t]
    \centering
    \includegraphics[width=\columnwidth]{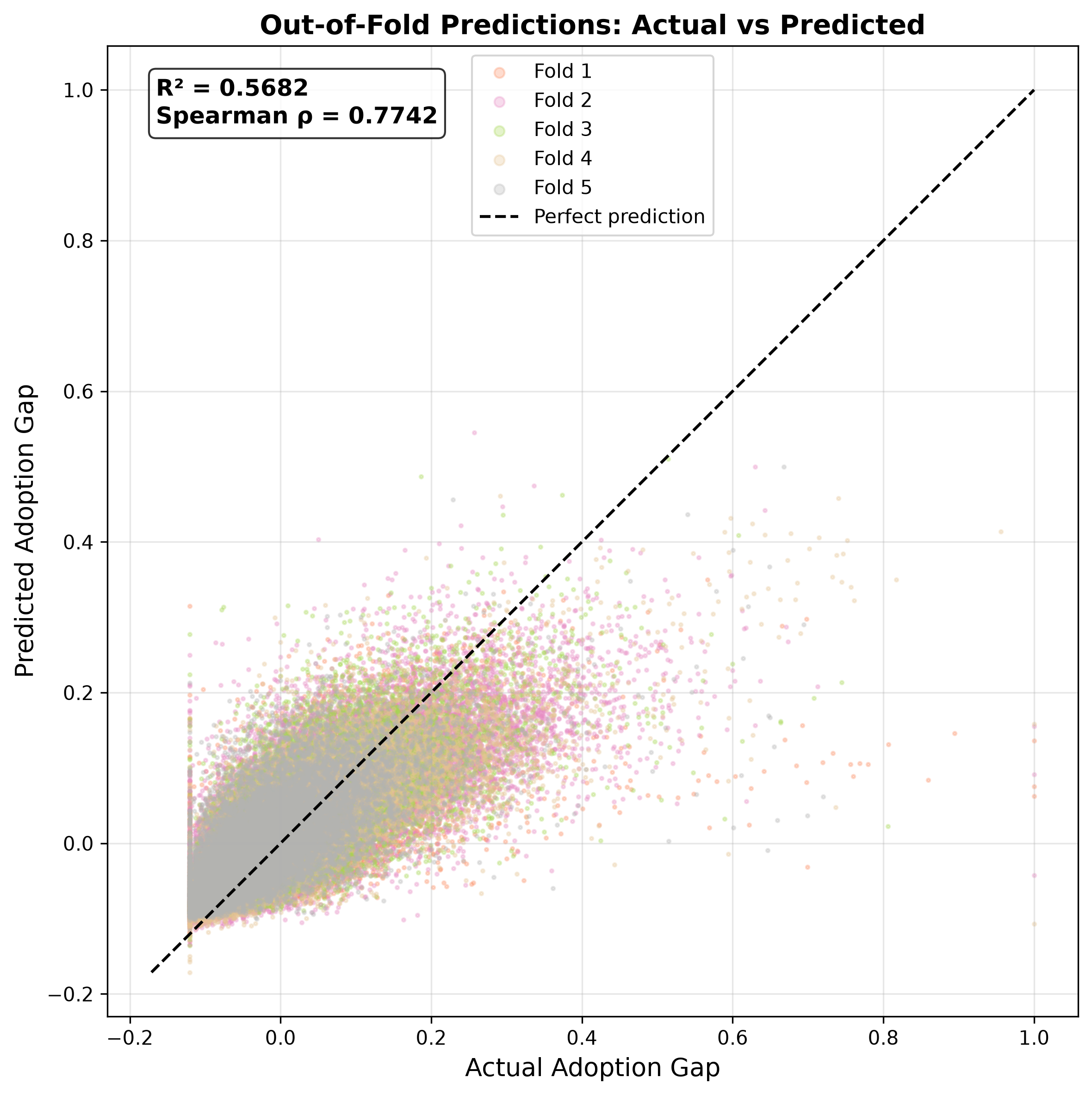}
    \caption{Actual versus predicted adoption gaps (out-of-fold predictions, $n = 83{,}359$). Pooled $R^2 = 0.568$, $\rho = 0.774$. Pooled values exceed the fold-mean ($R^2 = 0.533$) due to between-fold mean differences.}
    \label{fig:actual_vs_predicted}
\end{figure}

Fig.~\ref{fig:actual_vs_predicted} shows actual versus predicted adoption gaps from out-of-fold predictions. Moderate heteroscedasticity at extreme values is consistent with the reduced $R^2$ on the underserved subset.

\subsection{Temporal Stability Check}

Table~\ref{tab:temporal} summarizes the temporal stability check. National median broadband adoption rose from 78.7\% (ACS 2017) to 89.4\% (ACS 2022), a 10.7~pp level shift. The 2017-trained model achieves Spearman $\rho = 0.784$ on 2022 data, suggesting that the relative ranking of tracts by adoption gap is temporally stable (noting that hyperparameters were tuned on 2022 data, which may inflate this estimate). Temporal $\rho$ and within-vintage spatial-CV $\rho$ (0.763) are not directly comparable: the temporal model trains on all regions and faces no spatial extrapolation, while spatial CV forces regional holdout. Raw $R^2 = 0.187$ is depressed by the level shift; after linear recalibration, $R^2 = 0.598$ (an in-sample figure). Absolute screening capture drops from 38.0\% (within-vintage) to 23.3\% under temporal transfer, but the gain over income-only persists: $+3.2$~pp (23.3\% vs.\ 20.1\%).

\begin{table}[t]
\centering
\caption{Temporal Stability Check: ACS 2017$\to$2022}
\label{tab:temporal}
\begin{tabular}{lcc}
\toprule
\textbf{Metric} & \textbf{Within-vintage} & \textbf{Temporal} \\
\midrule
Spearman $\rho$ & 0.763 & \textbf{0.784} \\
$R^2$ (raw) & 0.533 & 0.187 \\
$R^2$ (recalibrated) & --- & 0.598 \\
Screening gain (pp) & +2.8 & +3.2 \\
\bottomrule
\end{tabular}
\end{table}

\section{Discussion}

\subsection{Key Findings}

The joint prominence of income and education features indicates that communities with both low income \textit{and} low education are associated with compounding digital exclusion, while communities with only one deficit often maintain adequate adoption. This interaction effect is consistent with why income-only screening is suboptimal: it cannot distinguish between low-income communities where education or age-related factors contribute most to the predicted gap versus those where affordability factors dominate.

Labor force participation emerges as the seventh most important feature and participates in strong interactions with income. This is consistent with the possibility that employment serves as a mediating pathway: labor force participation is associated with digital skills exposure and financial resources for subscriptions. The elderly--renter interaction identifies a population that may lack both digital literacy and housing stability, making them difficult to reach through conventional infrastructure programs.

The three exploratory factor profiles (Well-Connected, Affordability-Limited, and Rural--Elderly) offer a hypothesis-generating taxonomy whose practical value lies in directing different investigative lenses to tracts with different factor compositions. Profile~2 tracts (21K, predominantly urban) are characterized by income and education deficit patterns suggesting that demand-side subsidies and digital skills training may warrant investigation. Profile~3 tracts (13K, 99\% rural) are characterized by infrastructure-proxy and age-related patterns suggesting that supply-side deployment combined with age-appropriate digital literacy programs may be appropriate. Validating these intervention matches requires pilot programs with causal evaluation designs.

\subsection{Comparison with Prior Approaches}

Our model substantially outperforms the linear regression approach of Zahnd et al.~\cite{zahnd2022geographic}, which used similar census-tract data but reported only regression coefficients without predictive performance metrics. The $R^2 = 0.533$ achieved under spatial CV, a conservative evaluation protocol, demonstrates that non-linear methods capture substantial predictive signal beyond what linear models extract ($R^2 \leq 0.380$ for Ridge, the best linear model).

Compared to Nabi et al.~\cite{nabi2024red}, who achieved AUC $> 0.98$ for broadband map quality auditing, our lower $R^2$ reflects the fundamentally more difficult task of predicting adoption from socioeconomic variables alone. Map quality auditing involves detecting inconsistencies in reported data, while adoption prediction requires capturing complex behavioral and economic dynamics from proxy features.

\subsection{Policy Implications}

The improvement in screening precision from ML-based methods over income-only heuristics ($+2.8$~pp, $p < 0.002$) has implications for the \$42.45 billion BEAD program. More importantly, the SHAP-based factor decomposition enables \textit{differentiated} investigation, telling policymakers not just \textit{where} to investigate but \textit{what type} of factors are most strongly associated with each tract's predicted gap:

\begin{itemize}
    \item \textbf{Profile 2 (Affordability-Limited Severe):} These 21,116 tracts are characterized by income and education deficit patterns. With the ACP's June 2024 termination removing the largest affordability intervention, this cohort is particularly relevant for policy attention. Demand-side subsidies, income-adjusted pricing, and workforce-linked digital skills training may warrant investigation.
    \item \textbf{Profile 3 (Rural--Elderly):} These 12,826 tracts are characterized by infrastructure-proxy and age-related patterns. BEAD's infrastructure focus is well-suited to these areas, but the digital literacy component would require complementary Digital Equity Act funding.
\end{itemize}

The framework provides per-tract explanations that could inform state-level BEAD action plans by identifying which factor group is most strongly associated with each tract's predicted gap, subject to causal validation through pilot programs.

\subsection{Scope and Use-Case Rationale}

Because the ACS provides tract-level adoption rates directly, a natural question is why a predictive model is needed. The primary value is the per-tract factor decomposition: the adoption rate reveals \textit{that} a gap exists, but SHAP reveals which feature groups are most strongly associated with the predicted gap. Two potential extensions, small-area estimation (pooling information to stabilize noisy ACS estimates for low-population tracts) and nowcasting (training on a released vintage and recalibrating for current-year predictions), are plausible but not evaluated here and remain future directions. Note that all tracts enter the loss unweighted despite ACS margins of error varying by an order of magnitude, and predictors and outcome derive from the same survey instrument; common-source measurement error may inflate within-vintage performance.

\subsection{Methodological Considerations}

\textbf{Spatial cross-validation.} Standard random CV would likely yield $R^2 > 0.7$, but this overstates generalization by allowing spatially proximate tracts in both training and test sets. Our spatial CV provides a conservative estimate directly relevant to predicting gaps in understudied areas.

\textbf{SHAP faithfulness.} TreeSHAP provides exact Shapley values satisfying efficiency, symmetry, and null player properties~\cite{lundberg2020local}, a stronger guarantee than approximation methods. However, SHAP attribution among correlated features (income, poverty, education) is not uniquely identified~\cite{kumar2020problems}; the factor-profile labels should be interpreted as associated patterns, not as identified causal pathways.

\textbf{Non-trivial framework design.} Although the individual components (LightGBM, TreeSHAP, $k$-means) are established methods, the integrated framework requires several non-trivial design decisions: (1)~careful exclusion of 23 leaky internet-subscription features that would inflate performance but provide no policy-actionable signal; (2)~spatial CV with geographically coherent folds that prevents spatial information leakage; (3)~policy-category-aware SHAP aggregation that maps 65 features into interpretable factor categories aligned with existing federal program structures; and (4)~profile-to-investigation matching that translates SHAP factor compositions into differentiated policy hypotheses.

\subsection{Limitations}

We identify nine limitations. First, SHAP values reflect \textit{predictive} associations, not causal relationships; the framework identifies associated factors that suggest, but do not confirm, intervention targets requiring validation through randomized pilots or quasi-experimental evaluation.

Second, the ${\sim}$2\% of tracts excluded for insufficient ACS responses are disproportionately small, remote, and potentially tribal; these are plausibly among the most underserved and their absence limits coverage.

Third, the ACS five-year estimates (2018--2022) blend pre- and post-pandemic patterns. The Affordable Connectivity Program (ACP) ended in June 2024; its termination is a major affordability shock not reflected in the training data. BEAD's June 2025 restructuring shifted toward lowest-cost deployment, reducing alignment with demand-side diagnosis.

Fourth, the temporal crosswalk uses area-weighted aggregation; 41.9\% of 2010 tracts were split or merged, introducing interpolation noise.

Fifth, the model explains 53\% of gap variance; unmeasured factors (ISP competition, local policies, community anchor institutions) contribute to the residual. Both the target variable and most predictors derive from the same ACS survey vintage, so common-source measurement error may inflate explained variance.

Sixth, state-held-out CV reveals high variance ($\sigma = 0.105$); small, demographically distinct states (e.g., Vermont, $R^2 = 0.058$, $n = 192$) are poorly served by the national model.

Seventh, SHAP attribution among correlated features is not uniquely identified~\cite{kumar2020problems}; the factor-profile labels should be interpreted as associated patterns.

Eighth, the model includes six race/ethnicity composition features and proposes using outputs to prioritize public investment, yet no subgroup error audit has been performed; error rates may differ systematically across racial compositions, and a fairness evaluation is needed before deployment.

Ninth, all tracts enter the loss unweighted despite ACS margins of error varying by an order of magnitude; tracts with large MOE may exert disproportionate influence on the model.

\section{Conclusion}

We present an explainable ML framework for profiling broadband adoption disparities across 83,359 U.S. census tracts. LightGBM achieves $R^2 = 0.525$ under state-held-out CV ($R^2 = 0.533$ under spatial CV); a temporal stability check with zero survey-year overlap (ACS 2017$\to$2022) suggests ranking stability ($\rho = 0.784$, though hyperparameters were tuned on the 2022 vintage). TreeSHAP analysis reveals that income and education features collectively carry the largest attribution. SHAP-based clustering identifies three exploratory factor profiles (Well-Connected Moderate, Affordability-Limited Severe, and Rural--Elderly) with distinct factor compositions. The screening advantage over income-only heuristics is modest ($+2.8$~pp within-vintage, $+3.2$~pp temporal), but the primary value lies in the per-tract factor decomposition that identifies which feature groups are most strongly associated with each tract's predicted gap.

Future work should pursue several directions. First, incorporating FCC Broadband Data Collection availability data would enable separating infrastructure from other rural factors, directly addressing the proxy limitation of the Rural--Elderly profile. Second, temporal analysis using multiple ACS waves would enable studying how adoption gaps evolve and respond to interventions. Third, causal inference methods (e.g., regression discontinuity around subsidy eligibility thresholds, or difference-in-differences around BEAD deployment) would validate whether SHAP-identified associated factors are indeed causal levers. Fourth, testing the small-area estimation use case via MOE-stratified evaluation would establish whether the model improves upon raw ACS estimates for low-population tracts. Fifth, a subgroup fairness audit across racial compositions is needed before deployment. Sixth, integration with actual BEAD deployment data as it becomes available would enable model updating, creating a feedback loop between prediction and policy evaluation.

\section*{Data Availability}
All data used in this study are publicly available from the U.S. Census Bureau (American Community Survey 2022 five-year estimates; County Business Patterns 2022; Gazetteer files 2023).

\bibliographystyle{IEEEtran}

\begin{thebibliography}{10}
\providecommand{\url}[1]{#1}
\csname url@samestyle\endcsname
\providecommand{\newblock}{\relax}
\providecommand{\bibinfo}[2]{#2}
\providecommand{\BIBentrySTDinterwordspacing}{\spaceskip=0pt\relax}
\providecommand{\BIBentryALTinterwordstretchfactor}{4}
\providecommand{\BIBentryALTinterwordspacing}{\spaceskip=\fontdimen2\font plus
\BIBentryALTinterwordstretchfactor\fontdimen3\font minus
  \fontdimen4\font\relax}
\providecommand{\BIBforeignlanguage}[2]{{%
\expandafter\ifx\csname l@#1\endcsname\relax
\typeout{** WARNING: IEEEtran.bst: No hyphenation pattern has been}%
\typeout{** loaded for the language `#1'. Using the pattern for}%
\typeout{** the default language instead.}%
\else
\language=\csname l@#1\endcsname
\fi
#2}}
\providecommand{\BIBdecl}{\relax}
\BIBdecl

\bibitem{lai2021revisiting}
J.~Lai and N.~O. Widmar, ``Revisiting the digital divide in the {COVID-19}
  era,'' \emph{Applied Economic Perspectives and Policy}, vol.~43, no.~1, pp.
  458--464, 2021.

\bibitem{horrigan2010broadband}
\BIBentryALTinterwordspacing
J.~B. Horrigan, ``Broadband adoption and use in {America},'' Federal
  Communications Commission, OBI Working Paper Series~1, 2010. [Online].
  Available:
  \url{https://transition.fcc.gov/DiversityFAC/032410/consumer-survey-horrigan.pdf}
\BIBentrySTDinterwordspacing

\bibitem{iija2021}
{117th Congress}, ``Infrastructure investment and jobs act,'' Public Law
  117-58, 135 Stat.\ 429, 2021, division F: Broadband provisions, including
  BEAD (\$42.45B), Digital Equity Act (\$2.75B), Middle Mile (\$1B), Tribal
  Broadband (\$2B).

\bibitem{ntia2023bead}
\BIBentryALTinterwordspacing
{National Telecommunications and Information Administration}, ``{BEAD}
  allocation methodology,'' 2023. [Online]. Available:
  \url{https://broadbandusa.ntia.gov/bead-allocation-methodology}
\BIBentrySTDinterwordspacing

\bibitem{crs2025bead}
\BIBentryALTinterwordspacing
L.~Zhu, ``The broadband equity, access, and deployment ({BEAD}) program: Issues
  for the 119th congress,'' Congressional Research Service, Tech. Rep. R48666,
  2025. [Online]. Available: \url{https://www.congress.gov/crs-product/R48666}
\BIBentrySTDinterwordspacing

\bibitem{lundberg2020local}
S.~M. Lundberg, G.~Erion, H.~Chen, A.~DeGrave, J.~M. Prutkin, B.~Nair, R.~Katz,
  J.~Himmelfarb, N.~Bansal, and S.-I. Lee, ``From local explanations to global
  understanding with explainable {AI} for trees,'' \emph{Nature Machine
  Intelligence}, vol.~2, no.~1, pp. 56--67, 2020.

\bibitem{vandijk2006digital}
J.~A. G.~M. van Dijk, ``Digital divide research, achievements and
  shortcomings,'' \emph{Poetics}, vol.~34, no. 4--5, pp. 221--235, 2006.

\bibitem{vandijk2020digital}
------, \emph{The Digital Divide}.\hskip 1em plus 0.5em minus 0.4em\relax
  Cambridge, UK: Polity Press, 2020.

\bibitem{blank2014dimensions}
G.~Blank and D.~Groselj, ``Dimensions of internet use: Amount, variety, and
  types,'' \emph{Information, Communication \& Society}, vol.~17, no.~4, pp.
  417--435, 2014.

\bibitem{grubesic2006spatial}
T.~H. Grubesic, ``A spatial taxonomy of broadband regions in the {United
  States},'' \emph{Information Economics and Policy}, vol.~18, no.~4, pp.
  423--448, 2006.

\bibitem{grubesic2012broadband}
------, ``The {U.S.} national broadband map: Data limitations and
  implications,'' \emph{Telecommunications Policy}, vol.~36, no.~2, pp.
  113--126, 2012.

\bibitem{salemink2017rural}
K.~Salemink, D.~Strijker, and G.~Bosworth, ``Rural development in the digital
  age: A systematic literature review on unequal {ICT} availability, adoption,
  and use in rural areas,'' \emph{Journal of Rural Studies}, vol.~54, pp.
  360--371, 2017.

\bibitem{prieger2013broadband}
J.~E. Prieger, ``The broadband digital divide and the economic benefits of
  mobile broadband for rural areas,'' \emph{Telecommunications Policy},
  vol.~37, no. 6--7, pp. 483--502, 2013.

\bibitem{perrin2019smartphones}
\BIBentryALTinterwordspacing
A.~Perrin and E.~Turner, ``Smartphones help blacks, hispanics bridge some---but
  not all---digital gaps with whites,'' Pew Research Center, Tech. Rep., 2019.
  [Online]. Available:
  \url{https://www.pewresearch.org/internet/2019/08/20/smartphones-help-blacks-hispanics-bridge-some-but-not-all-digital-gaps/}
\BIBentrySTDinterwordspacing

\bibitem{whitacre2014broadband}
B.~Whitacre, R.~Gallardo, and S.~Strover, ``Broadband's contribution to
  economic growth in rural areas: Moving toward a causal relationship,''
  \emph{Telecommunications Policy}, vol.~38, no.~11, pp. 1011--1023, 2014.

\bibitem{whitacre2014rural}
------, ``Does rural broadband impact jobs and income? {Evidence} from spatial
  and first-differenced regressions,'' \emph{The Annals of Regional Science},
  vol.~53, no.~3, pp. 649--670, 2014.

\bibitem{whitacre2015infrastructure}
B.~Whitacre, S.~Strover, and R.~Gallardo, ``How much does broadband
  infrastructure matter? {Decomposing} the metro--non-metro adoption gap with
  the help of the national broadband map,'' \emph{Government Information
  Quarterly}, vol.~32, no.~3, pp. 261--269, 2015.

\bibitem{gallardo2022ddi}
\BIBentryALTinterwordspacing
R.~Gallardo, ``The state of the digital divide in the {United States},'' Purdue
  Center for Regional Development, 2022. [Online]. Available:
  \url{https://pcrd.purdue.edu/the-state-of-the-digital-divide-in-the-united-states/}
\BIBentrySTDinterwordspacing

\bibitem{oughton2021predicting}
E.~J. Oughton and J.~Mathur, ``Predicting cell phone adoption metrics using
  machine learning and satellite imagery,'' \emph{Telematics and Informatics},
  vol.~62, p. 101622, 2021.

\bibitem{singleton2020mapping}
A.~Singleton, A.~Alexiou, and R.~Savani, ``Mapping the geodemographics of
  digital inequality in {Great Britain}: An integration of machine learning
  into small area estimation,'' \emph{Computers, Environment and Urban
  Systems}, vol.~82, p. 101486, 2020.

\bibitem{zahnd2022geographic}
W.~E. Zahnd, N.~Bell, and A.~E. Larson, ``Geographic, racial/ethnic, and
  socioeconomic inequities in broadband access,'' \emph{The Journal of Rural
  Health}, vol.~38, no.~3, pp. 519--530, 2022.

\bibitem{paul2023decoding}
U.~Paul, V.~Gunasekaran, J.~Liu, T.~N. Narechania, A.~Gupta, and E.~Belding,
  ``Decoding the divide: Analyzing disparities in broadband plans offered by
  major {US} {ISPs},'' in \emph{Proceedings of the ACM SIGCOMM 2023
  Conference}, 2023.

\bibitem{nabi2024red}
S.~T. Nabi, Z.~Wen, B.~Ritter, and S.~Hasan, ``Red is sus: Automated
  identification of low-quality service availability claims in the {US}
  national broadband map,'' in \emph{Proceedings of the 2024 ACM Internet
  Measurement Conference (IMC)}, 2024.

\bibitem{agarwal2024rural}
A.~Agarwal and C.~Canfield, ``Analysis of rural broadband adoption dynamics: A
  theory-driven agent-based model,'' \emph{PLOS ONE}, vol.~19, no.~6, p.
  e0302146, 2024.

\bibitem{lundberg2017unified}
S.~M. Lundberg and S.-I. Lee, ``A unified approach to interpreting model
  predictions,'' in \emph{Advances in Neural Information Processing Systems 30
  (NeurIPS)}, 2017, pp. 4765--4774.

\bibitem{rudin2019stop}
C.~Rudin, ``Stop explaining black box machine learning models for high stakes
  decisions and use interpretable models instead,'' \emph{Nature Machine
  Intelligence}, vol.~1, no.~5, pp. 206--215, 2019.

\bibitem{athey2019machine}
S.~Athey and G.~W. Imbens, ``Machine learning methods that economists should
  know about,'' \emph{Annual Review of Economics}, vol.~11, pp. 685--725, 2019.

\bibitem{lundberg2018explainable}
S.~M. Lundberg, B.~Nair, M.~S. Vavilala, M.~Horibe, M.~J. Eisses, T.~Adams,
  D.~E. Liston, D.~K.-W. Low, S.-F. Newman, J.~Kim, and S.-I. Lee,
  ``Explainable machine-learning predictions for the prevention of hypoxaemia
  during surgery,'' \emph{Nature Biomedical Engineering}, vol.~2, no.~10, pp.
  749--760, 2018.

\bibitem{wagner2022shap}
F.~Wagner, N.~Milojevic-Dupont, L.~Franken, A.~Zekar, B.~Thies, N.~Koch, and
  F.~Creutzig, ``Using explainable machine learning to understand how urban
  form shapes sustainable mobility,'' \emph{Transportation Research Part D:
  Transport and Environment}, vol. 111, p. 103442, 2022.

\bibitem{digitalequityact2021}
{117th Congress}, ``Digital equity act of 2021,'' Division F, Title III,
  Sections 60302--60307, Pub.\ L.\ No.\ 117-58, 2021, appropriated \$2.75B for
  State Digital Equity Planning Grants, Capacity Grants, and Competitive
  Grants.

\bibitem{ke2017lightgbm}
G.~Ke, Q.~Meng, T.~Finley, T.~Wang, W.~Chen, W.~Ma, Q.~Ye, and T.-Y. Liu,
  ``{LightGBM}: A highly efficient gradient boosting decision tree,'' in
  \emph{Advances in Neural Information Processing Systems 30 (NeurIPS)}, 2017,
  pp. 3146--3154.

\bibitem{akiba2019optuna}
T.~Akiba, S.~Sano, T.~Yanase, T.~Ohta, and M.~Koyama, ``Optuna: A
  next-generation hyperparameter optimization framework,'' in \emph{Proceedings
  of the 25th ACM SIGKDD International Conference on Knowledge Discovery and
  Data Mining}, 2019, pp. 2623--2631.

\bibitem{roberts2017cross}
D.~R. Roberts, V.~Bahn, S.~Ciuti, M.~S. Boyce, J.~Elith, G.~Guillera-Arroita,
  S.~Hauenstein, J.~J. Lahoz-Monfort, B.~Schr\"{o}der, W.~Thuiller, D.~I.
  Warton, B.~A. Wintle, F.~Hartig, and C.~F. Dormann, ``Cross-validation
  strategies for data with temporal, spatial, hierarchical, or phylogenetic
  structure,'' \emph{Ecography}, vol.~40, no.~8, pp. 913--929, 2017.

\bibitem{chen2016xgboost}
T.~Chen and C.~Guestrin, ``{XGBoost}: A scalable tree boosting system,'' in
  \emph{Proceedings of the 22nd ACM SIGKDD International Conference on
  Knowledge Discovery and Data Mining}, 2016, pp. 785--794.

\bibitem{kumar2020problems}
I.~E. Kumar, S.~Venkatasubramanian, C.~Scheidegger, and S.~Friedler, ``Problems
  with {Shapley}-value-based explanations as feature importance measures,'' in
  \emph{Proceedings of the 37th International Conference on Machine Learning
  (ICML)}, 2020, pp. 5491--5500.

\end{thebibliography}

\end{document}